\documentclass[10pt,twocolumn,letterpaper]{article}

\usepackage{iccv}
\usepackage{times}
\usepackage{epsfig}
\usepackage{graphicx}
\usepackage{amsmath}
\usepackage{amssymb}
\usepackage{gensymb}
\usepackage{abbrevs}
\usepackage{comment}
\newabbrev\modelDA{CCSA}
\newabbrev\modelGA{CCSA}

\usepackage[breaklinks=true,bookmarks=false]{hyperref}

\iccvfinalcopy 

\def\iccvPaperID{****} 

\ificcvfinal\fi
\begin{document}

\title{Unified Deep Supervised Domain Adaptation and Generalization}

\author{Saeid Motiian $\quad$ Marco Piccirilli $\quad$ Donald A. Adjeroh $\quad$ Gianfranco Doretto\\
West Virginia University\\
Morgantown, WV 26508\\
{\tt\small \{samotiian, mpiccir1, daadjeroh, gidoretto\}@mix.wvu.edu}}

\maketitle

\begin{abstract}
This work provides a unified framework for addressing the problem of visual supervised domain adaptation and generalization with deep models. The main idea is to exploit the Siamese architecture to learn an embedding subspace that is discriminative, and where mapped visual domains are semantically aligned and yet maximally separated. The supervised setting becomes attractive especially when only few target data samples need to be labeled. In this scenario, alignment and separation of semantic probability distributions is difficult because of the lack of data. We found that by reverting to point-wise surrogates of distribution distances and similarities provides an effective solution. In addition, the approach has a high ``speed'' of adaptation, which requires an extremely low number of labeled target training samples, even one per category can be effective. The approach is extended to domain generalization. For both applications the experiments show very promising results.
\end{abstract}

\vspace{-4mm}
\section{Introduction} \label{sec-Introduction}

Many computer vision applications require enough labeled data (\emph{target data}) for training visual classifiers to address a specific task at hand. Whenever target data is either not available, or it is expensive to collect and/or label it, the typical approach is to use available datasets (\emph{source data}), representative of a closely related task. Since this practice is known for leading to suboptimal performance, techniques such as \emph{domain adaptation}~\cite{blitzer2006domain} and/or \emph{domain generalization}~\cite{blanchard2011generalizing} have been developed to address the issue. Domain adaptation methods require target data, whereas domain generalization methods do not. Domain adaptation can be either \emph{supervised}~\cite{tzengHDS15iccv,koniusz2016domain}, \emph{unsupervised}~\cite{liu2016coupled,liu2016coupled,Tzeng_2017_CVPR}, or \emph{semi-supervised}~\cite{gong2012geodesic,GuoX12,Yao_2015_CVPR}. Unsupervised domain adaptation (UDA) is attractive because it does not require target data to be labeled. Conversely, supervised domain adaptation (SDA) requires labeled target data. 

UDA expects large amounts of target data in order to be effective, and this is emphasized even more when using deep models. Moreover, given the same amount of target data, SDA typically outperforms UDA, as we will later explain. Therefore, especially when target data is \emph{scarce}, it is more attractive to use SDA, also because limited amounts of target data are likely to not be very expensive to label. 

In the absence of target data, domain generalization (DG) exploits several cheaply available datasets (sources), representing different specific but closely related tasks. It then attempts to learn by combining data sources in a way that produces visual classifiers that are less sensitive to the specific target data that will need to be processed.

In this work, we introduce a supervised approach for visual recognition that can be used for both SDA and DG. The SDA approach requires very few labeled target samples per category in training. Indeed, even one sample can significantly increase performance, and a few others bring it closer to a peak, showing a remarkable ``speed'' of adaptation. Moreover, the approach is also robust to adapting to categories that have no target labeled samples. Although domain adaptation and generalization are closely related, adaptation techniques are not directly applied to DG, and viceversa. However, we show that by making simple changes to our proposed training loss function, and by maintaining the same architecture, our SDA approach very effectively extends to DG.

Using basic principles, we analyze how visual classification is extended to handle UDA by aligning a \emph{source domain} distribution to a \emph{target domain} distribution to make the classifier domain invariant. This leads to observing that SDA approaches improve upon UDA by making the alignment \emph{semantic}, because they can ensure the alignment of semantically equivalent distributions from different domains. However, we go one step ahead by suggesting that semantic distribution separation should further increase performance, and this leads to the introduction of a \emph{classification and contrastive semantic alignment} (CCSA) loss.

\begin{figure*}[t!]
\begin{center}
   \includegraphics[width=0.9\linewidth]{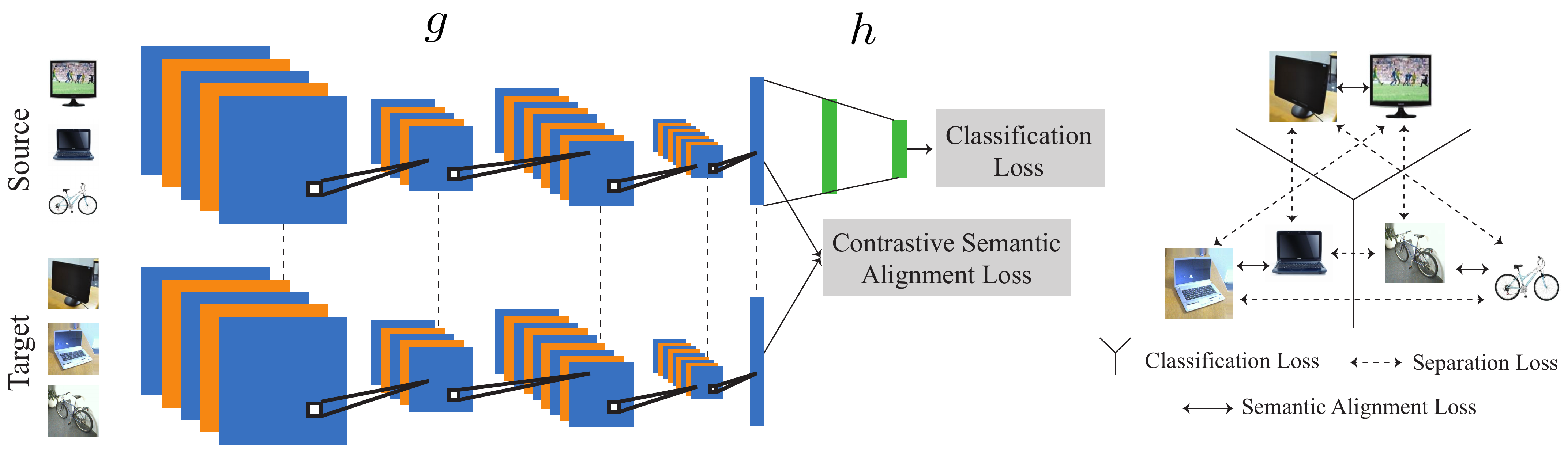}
\end{center}
   \caption{\textbf{Deep supervised domain adaptation.} In training, the semantic alignment loss minimizes the distance between samples from different domains
but the same class label and the separation loss maximizes the distance between samples from different domains
and class labels. At the same time, the classification loss guarantees high classification accuracy.}
\label{fig-model}
\end{figure*}

We deal with the limited size of target domain samples by observing that the CCSA loss relies on computing distances and similarities between distributions (as typically done in adaptation and generalization approaches). Those are difficult to represent with limited data. Thus, we revert to point-wise surrogates. The resulting approach turns out to be very effective as shown in the experimental section.


\begin{figure*}[t!]
\centering
\includegraphics[width=0.9\linewidth]{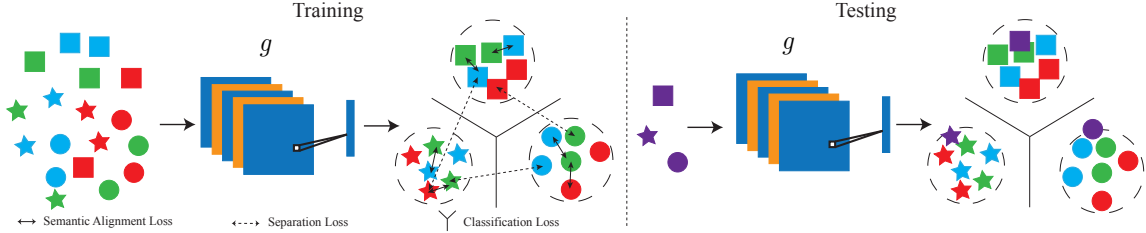}
\caption{\textbf{Deep domain generalization.} In training, the semantic alignment loss minimizes the distance between samples from different domains
but the same class label and the separation loss maximizes the distance between samples from different domains
and class labels. At the same time, the classification loss guarantees high classification accuracy. In testing, the embedding function embeds samples from unseen distributions to the domain invariant space and the prediction function classifies them (right). In this figure, different colors represent different domain distributions and different shapes represent different classes.}
\label{fig-domain_generalization}
\end{figure*}

\section{Related work}

\noindent \textbf{Domain adaptation.} Visual recognition algorithms are trained with data from a \emph{source domain}, and when they are tested on a \emph{target domain} with marginal distribution that differs from the one of the sources, we experience the visual domain adaptation (DA) problem (also known as dataset bias~\cite{ponce2006dataset,torralbaE11cvpr,tommasi2015deeper}, or covariate shift~\cite{shimodaira00jspi}), and observe a performance decrease. 

Traditional DA methods attempt to directly minimize the shift between source and target distributions. We divide them in three categories. The first one includes those that try to find a mapping between source and target distributions~\cite{saenkoKFD2010eccv,kulis2011you,gopalanLC11iccv,gong2012geodesic,fernandoHST13iccv,tommasi2016learning}. The second one seeks to find a shared latent space for source and target distributions~\cite{long2013transfer,baktashmotlaghHLS13iccv,muandet2013domain,ganin2014unsupervised,ganin2016domain,panTKY11tnn,motiian2016information}. The third one regularizes a classifier trained on a source distribution to work well on a target distribution~\cite{bergamo2010exploiting,aytar2011tabula,yang2007adapting,duan2009domain,becker2013non,daume2006domain}. UDA approaches fall in the first and second categories, while SDA methods could fall either in the second or third category or sometimes both. Recently, \cite{chenLX2014cvpr,motiian2016ECCV} have addressed UDA when an auxiliary data view~\cite{lapinHS2014nn,motiian2016information}, is available during training, which is beyond the scope of this work.

Here, we are interested in finding a shared subspace for source and target distributions. Among algorithms for subspace learning, Siamese networks~\cite{chopra2005learning} work well for different tasks~\cite{Donahue13decaf,Simonyan14c,kumar2016learning,varior2016siamese,chen2015deep}. Recently, Siamese networks have been used for domain adaptation. In~\cite{tzengHDS15iccv}, which is an SDA approach, unlabeled and sparsely labeled target domain data are used to optimize for domain invariance to facilitate domain transfer while using a soft label distribution matching loss. In~\cite{sun2016deep}, which is a UDA approach, unlabeled target data is used to learn a nonlinear transformation that aligns correlations of layer activations in deep neural networks. Some approaches went beyond the Siamase weight-sharing and used couple networks for DA. \cite{koniusz2016domain} uses two CNN streams, for source and target, fused at the classifier level. 
\cite{rozantsev2016beyond} uses a two-streams architecture, for source and target, with related but not shared weights. Here we use a Siamese network to learn an embedding such that samples from the same class are mapped as close as possible to each other. This semantic alignment objective is similar to other deep approaches, but unlike them, we explicitly model and introduce cross-domain class separation forces. Moreover, we do so with very few training samples, which makes the problem of characterizing distributions challenging, and this is why we propose to use point-wise surrogates.

\noindent \textbf{Domain generalization.} Domain generalization (DG) is a less investigated problem and is addressed in two ways. In the first one, all information from the training domains or datasets is aggregated to learn a shared invariant representation. Specifically, \cite{blanchard2011generalizing} pulls all of the training data together in one dataset, and learns a single SVM classifier. \cite{muandet2013domain} learns an invariant transformation by minimizing the dissimilarity across domains. \cite{ghifary2016scatter}, which can be used for SDA too, finds a representation that minimizes the mismatch between domains and maximizes the separability of data. \cite{ghifary2015domain} learns features that are robust to variations across domains.

The second approach to DG is to exploit all information from the training domains to train a classifier or regulate its weights~\cite{khosla2012undoing,FXRQ_iccv13,xuLNX14eccv,niu2015multi,niu2016exemplar}. 
Specifically, \cite{khosla2012undoing} adjusts the weights of the classifier to work well on an unseen dataset, and~\cite{xuLNX14eccv} fuses the score of exemplar classifiers given any test sample. While most works use the shallow models, here we approach DG as in the first way, and extend the proposed SDA approach by training a deep Siamese network to find a shared invariant representation where semantic alignment as well as separation are explicitly accounted for. To the best of our knowledge, \cite{ghifary2015domain} is the only DG approach using deep models, and our method is the first deep method that solves both adaptation and generalization.


\section{Supervised DA with Scarce Target Data} \label{sec-DA}


In this section we describe the model we propose to address supervised domain adaptation (SDA), and in the following Section~\ref{sec-Deep-DG} we extend it to address the domain generalization problem.
We are given a training dataset made of pairs $\mathcal{D}_s = \{ (x_i^s, y_i^s) \}_{i=1}^N$. The feature $x_i^s \in \mathcal{X}$ is a realization from a random variable $X^s$, and the label $y_i^s \in \mathcal{Y}$ is a realization from a random variable $Y$. In addition, we are also given the training data $\mathcal{D}_t = \{ (x_i^t, y_i^t) \}_{i=1}^M$, where $x_i^t \in \mathcal{X}$ is a realization from a random variable $X ^t$, and the labels $y_i^t \in \mathcal{Y}$. We assume that there is a \emph{covariate shift}~\cite{shimodaira00jspi} between $X^s$ and $X^t$, i.e., there is a difference between the probability distributions $p(X^s)$ and $p(X^t)$. We say that $X^s$ represents the \emph{source domain} and that $X^t$ represents the \emph{target domain}. Under this settings the goal is to learn a prediction function $f : \mathcal{X} \rightarrow \mathcal{Y}$ that during testing is going to perform well on data from the target domain. 

The problem formulated thus far is typically referred to as \emph{supervised domain adaptation}. In this work we are especially concerned with the version of this problem where only very few target labeled samples per class are available. We aim at handling cases where there is only one target labeled sample, and there can even be some classes with no target samples at all.

\subsection{Deep SDA} 
\label{sec-DA-basic}

In the absence of covariate shift a visual classifier $f$ is trained by minimizing a \emph{classification loss}
\begin{equation}
\mathcal{L}_C (f) = E[ \ell (f(X^s), Y) ] \; ,
\label{eq-basic}
\end{equation}
where $E[ \cdot ]$ denotes statistical expectation and $\ell$ could be any appropriate loss function (for example categorical cross-entropy for multi-class classification). When the distributions of  $X^s$ and $X^t$ are different, a deep model $f_s$ trained with $\mathcal{D}_s$ will have reduced performance on the target domain. Increasing it would be trivial by simply training a new model $f_t$ with data $\mathcal{D}_t$. However, $\mathcal{D}_t$ is small and deep models require large amounts of labeled data.

In general, $f$ could be modeled by the composition of two functions, i.e., $f = h \circ g$. Here $g : \mathcal{X} \rightarrow \mathcal{Z}$ would be an embedding from the input space $\mathcal{X}$ to a feature or embedding space $\mathcal{Z}$, and $h : \mathcal{Z} \rightarrow \mathcal{Y}$ would be a function for predicting from the feature space. With this notation we would have $f_s = h_s \circ g_s$ and $f_t = h_t \circ g_t$, and the SDA problem would be about finding the best approximation for $g_t$ and $h_t$, given the constraints on the available data.

The unsupervised DA paradigm (UDA) assumes that $\mathcal{D}_t$ does not have labels. In that case the typical approach assumes that $g_t = g_s = g$, and $f$ minimizes~\eqref{eq-basic}, while $g$ also minimizes 
\begin{equation}
\mathcal{L}_{CA} (g) = d(p(g(X^s)),p(g(X^t)))\; .
\label{eq-uda-confusion}
\end{equation}
The purpose of~\eqref{eq-uda-confusion} is to align the distributions of the features in the embedding space, mapped from the source and the target domains. $d$ is meant to be a metric between distributions that once aligned, they will no longer allow to tell whether a feature is coming from the source or the target domain. For that reason, we refer to~\eqref{eq-uda-confusion} as the \emph{confusion alignment} loss. A popular choice for $d$ is the Maximum Mean Discrepancy~\cite{grettonBRSS06nips}. In the embedding space $\mathcal{Z}$, features are assumed to be domain invariant. Therefore, UDA methods say that from the feature to the label space it is safe to assume that $h_t = h_s = h$. 

Since we are interested in visual recognition, the embedding function $g$ would be modeled by a convolutional neural network (CNN) with some initial convolutional layers, followed by some fully connected layers. In addition, the training architecture would have two streams, one for source and the other for target samples. Since $g_s = g_t = g$, the CNN parameters would be shared as in a Siamese architecture. In addition, the source stream would continue with additional fully connected layers for modeling $h$. See Figure~\ref{fig-model}.

\begin{figure*}[th!]
\centering
\includegraphics[width=0.33\linewidth]{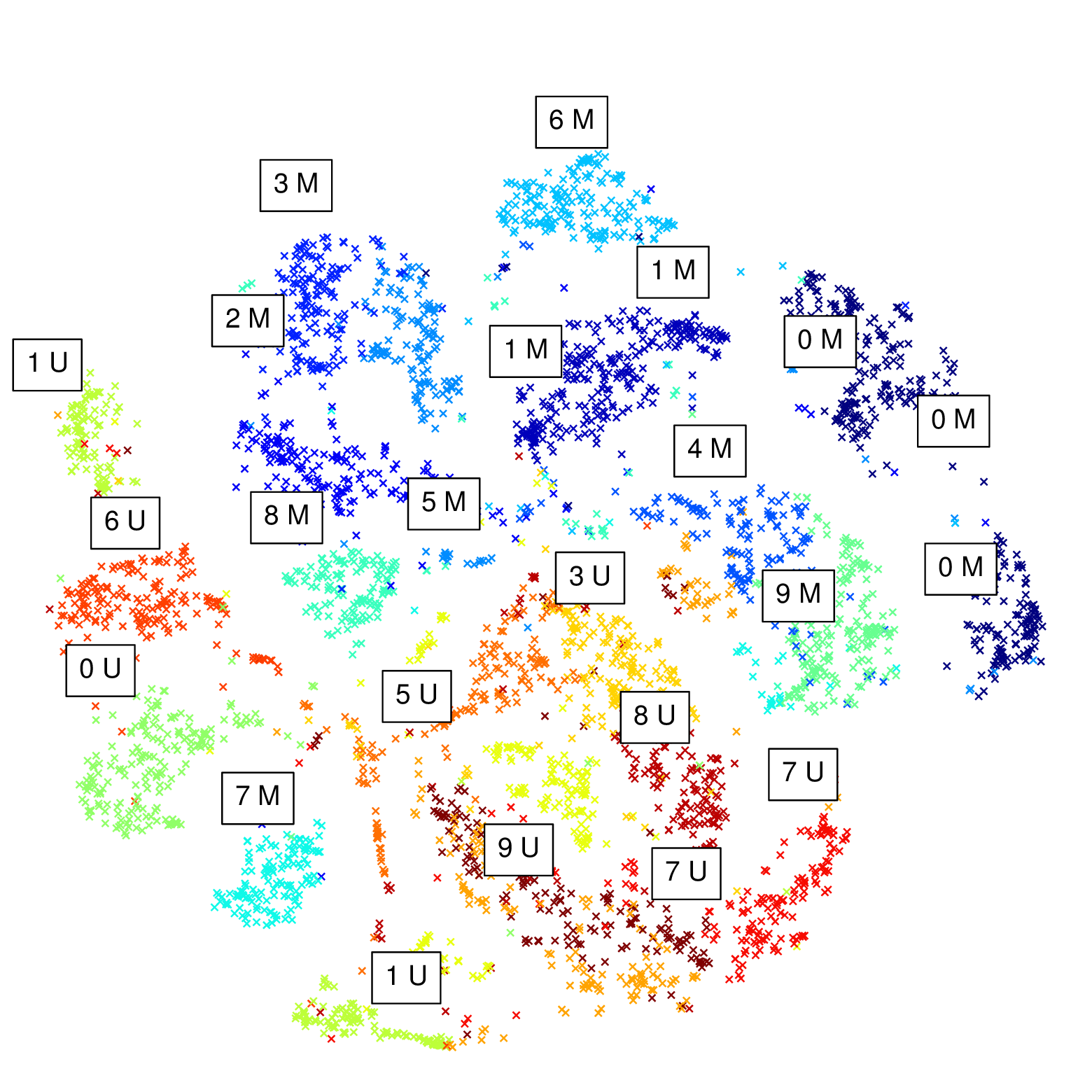}
\includegraphics[width=0.33\linewidth]{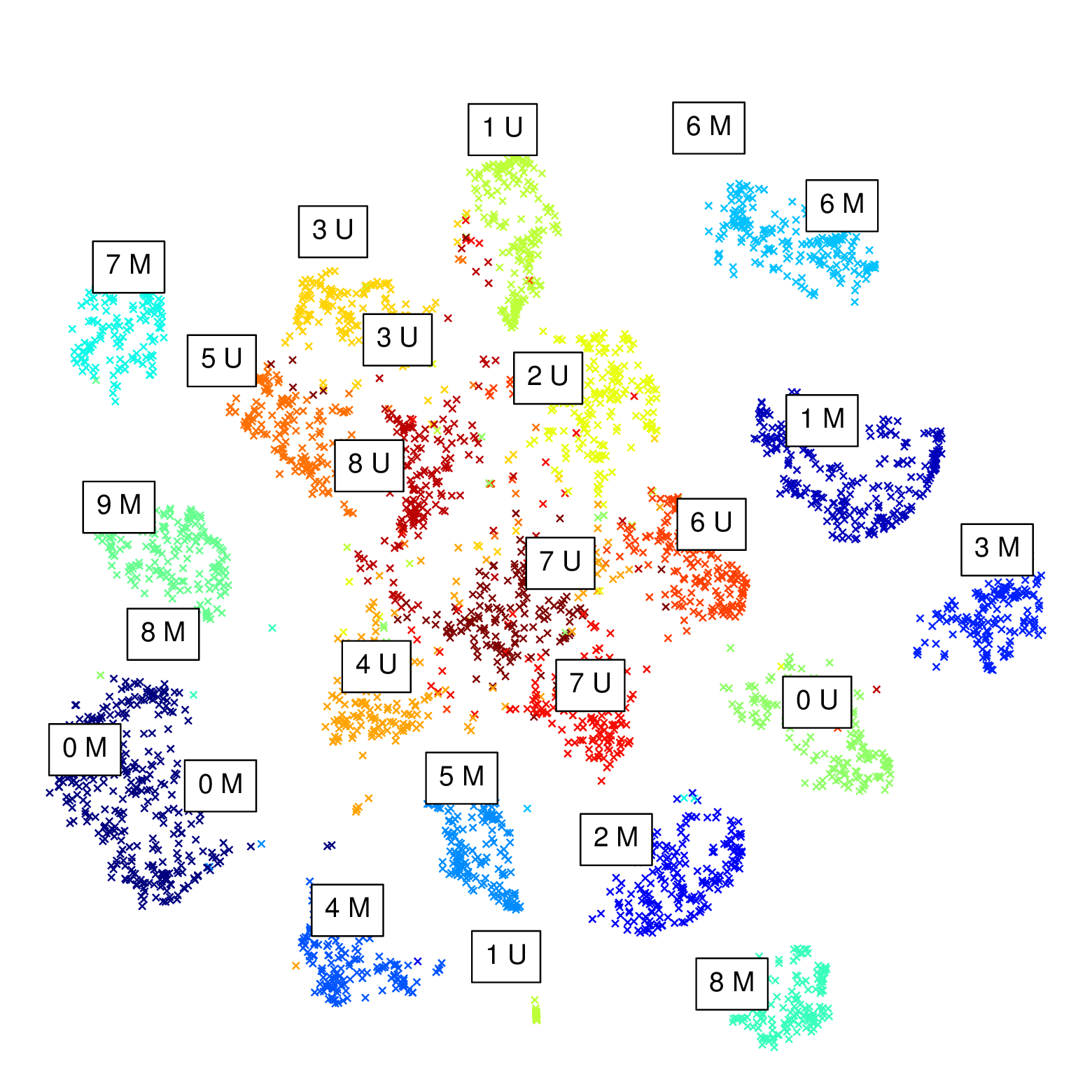}
\includegraphics[width=0.33\linewidth]{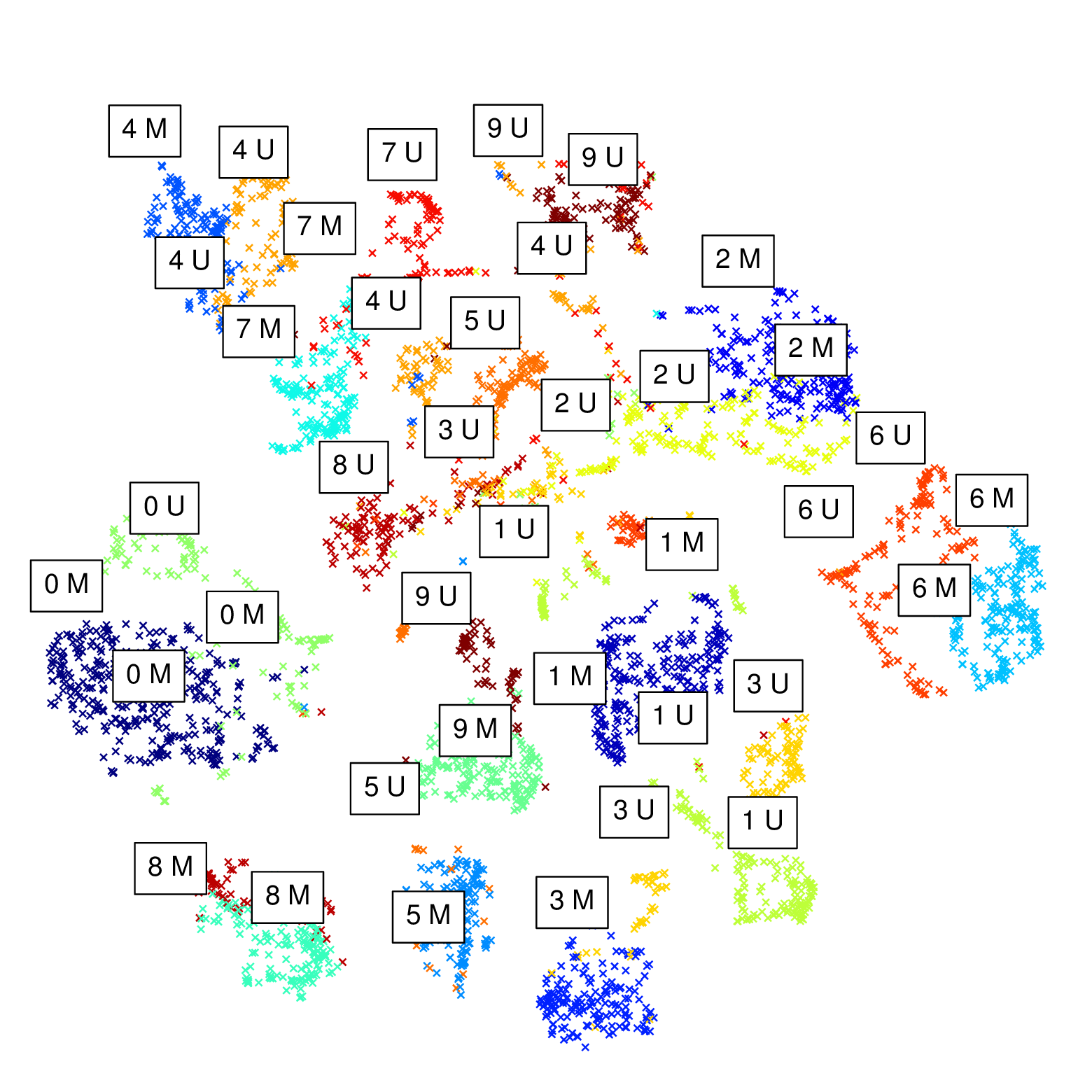}
\caption{\textbf{Visualization of the MNIST-USPS datasets.} {\bf Left}: 2D visualization of the row images of the MNIST-USPS datasets. The samples from the same class and different domains lie far from each other on the 2D subspace. {\bf Middle}: 2D visualization of the embedded images using our base model (without domain adaptation). The samples from the same class and different domains still lie far from each other on the 2D subspace. {\bf Right}: 2D visualization of the embedded images using our SDA model. The samples from the same class and different domains lie very close to each other on the 2D subspace. }
\label{fig-domain_embeded}
\end{figure*}

From the above discussion it is clear that in order to perform well, UDA needs to align effectively. This can happen only if distributions are represented by a sufficiently large dataset. Therefore, UDA approaches are in a position of weakness because we assume $\mathcal{D}_t$ to be small. Moreover, UDA approaches have also another intrinsic limitation, which is that even with perfect confusion alignment, there is no guarantee that samples from different domains but the same class label, would map nearby in the embedding space. This lack of \emph{semantic alignment} is a major source of performance reduction.

SDA approaches easily address the semantic alignment problem by replacing~\eqref{eq-uda-confusion} with
\begin{equation}
\mathcal{L}_{SA} (g) = \sum_{a=1}^C d(p(g(X_a^s)),p(g(X_a^t)))\; ,
\label{eq-sda-semantic-confusion}
\end{equation}
where $C$ is the number of class labels, and $X_a^s = X^s|\{Y=a\}$ and $X_a^t = X^t|\{Y=a\}$ are conditional random variables. $d$ instead is a suitable distance metric between the distributions of $X_a^s$ and $X_a^t$ in the embedding space. We refer to~\eqref{eq-sda-semantic-confusion} as the \emph{semantic alignment} loss, which clearly encourages samples from different domains but the same label, to map nearby in the embedding space.

While the analysis above clearly indicates why SDA provides superior performance than UDA, it also suggests that deep SDA approaches have not considered that greater performance could be achieved by encouraging class separation, meaning that samples from different domains and with different labels, should be mapped as far apart as possible in the embedding space. This idea means that, in principle, a semantic alignment less prone to errors should be achieved by adding to~\eqref{eq-sda-semantic-confusion} the following term
\begin{equation}
\mathcal{L}_{S} (g) = \sum_{a,b |a\neq b} k(p(g(X_a^s)),p(g(X_b^t))) \; ,
\label{eq-sda-separation}
\end{equation}
where $k$ is a suitable similarity metric between the distributions of $X_a^s$ and $X_b^t$ in the embedding space, which adds a penalty when the distributions $p(g(X_a^s))$ and $p(g(X_b^t))$ come close, since they would lead to lower classification accuracy. We refer to~\eqref{eq-sda-separation} as the \emph{separation} loss.

Finally, we suggest that SDA could be approached by learning a deep model $f=h \circ g$ such that
\begin{equation}
\mathcal{L}_{CCSA} (f) = \mathcal{L}_{C} (h\circ g) + \mathcal{L}_{SA} (g) + \mathcal{L}_{S} (g)\; .
\label{eq-sda-contrastive-semantic-alignment}
\end{equation}
We refer to~\eqref{eq-sda-contrastive-semantic-alignment} as the \emph{classification and contrastive semantic alignment} loss. This would allow to set $g_s = g_t = g$. The classification network $h$ is trained only with source data, so $h_s = h$. In addition, to improve performance on the target domain, $h_t$ could be obtained via fine-tuning based on the few samples in $\mathcal{D}_t$, i.e.,
\begin{equation}
h_t = \text{fine-tuning} (h | \mathcal{D}_t) \; .
\end{equation}
Note that the network architecture remains the one in Figure~\ref{fig-model}, only with a different loss, and training procedure.

\subsection{Handling Scarce Target Data} \label{sec-Deep-DA}

When the size of the labeled target training dataset $\mathcal{D}_t$ is very small, minimizing the loss~\eqref{eq-sda-contrastive-semantic-alignment} becomes a challenge. The problem is that the semantic alignment loss as well as the separation loss rely on computing distances and similarities between distributions, and those are very difficult to represent with as few as one data sample. 

Rather than attempting to characterize distributions with statistics that require enough data, because of the reduced size of $\mathcal{D}_t$, we compute the distance in the semantic alignment loss~\eqref{eq-sda-semantic-confusion} by computing average pairwise distances between points in the embedding space, i.e., we compute
\begin{equation}
d(p(g(X_a^s)),p(g(X_a^t))) = \sum_{i,j} d(g(x_i^s),g(x_j^t)) \; ,
\label{eq-semantic-distance}
\end{equation}
where it is assumed $y_i^s = y_j^t = a$. The strength of this approach is that it allows even a single labeled target sample to be paired with all the source samples, effectively trying to semantically align the entire source data with the few target data. Similarly, we compute the similarities in the separation loss~\eqref{eq-sda-separation} by computing average pairwise similarities between points in the embedding space, i.e., we compute
\begin{equation}
k(p(g(X_a^s)),p(g(X_b^t))) = \sum_{i,j} k(g(x_i^s),g(x_j^t)) \; ,
\label{eq-separation-similarity}
\end{equation}
where it is assumed that $y_i^s = a \neq y_j^t = b$.

Moreover, our implementation further assumes that
\begin{eqnarray}
d(g(x_i^s),g(x_j^t)) &=& \frac{1}{2} \| g(x_i^s) - g(x_j^t) \|^2 \; , \label{eq-distance-fobenius}
\\
k(g(x_i^s),g(x_j^t)) &=& \frac{1}{2} \max( 0, m - \| g(x_i^s) - g(x_j^t) \| )^2 \;\;\;\;\;\; \label{eq-similarity-fobenius}
\end{eqnarray}
where $\|\cdot\|$ denotes the Frobenius norm, and $m$ is the margin that specifies the separability in the embedding space. Note that with the choices outlined in~\eqref{eq-distance-fobenius} and~\eqref{eq-similarity-fobenius}, the loss $\mathcal{L}_{SA}(g)+\mathcal{L}_{S}(g)$ becomes the well known  contrastive loss as defined in~\cite{hadsell2006dimensionality}. Finally, to balance the classification versus the contrastive semantic alignment portion of the loss~\eqref{eq-sda-contrastive-semantic-alignment}, \eqref{eq-semantic-distance} and~\eqref{eq-separation-similarity} are normalized and weighted by $1-\gamma$ and~\eqref{eq-basic} by $\gamma$.


\begin{figure}[t]
\begin{center}
\includegraphics[width=0.48\linewidth]{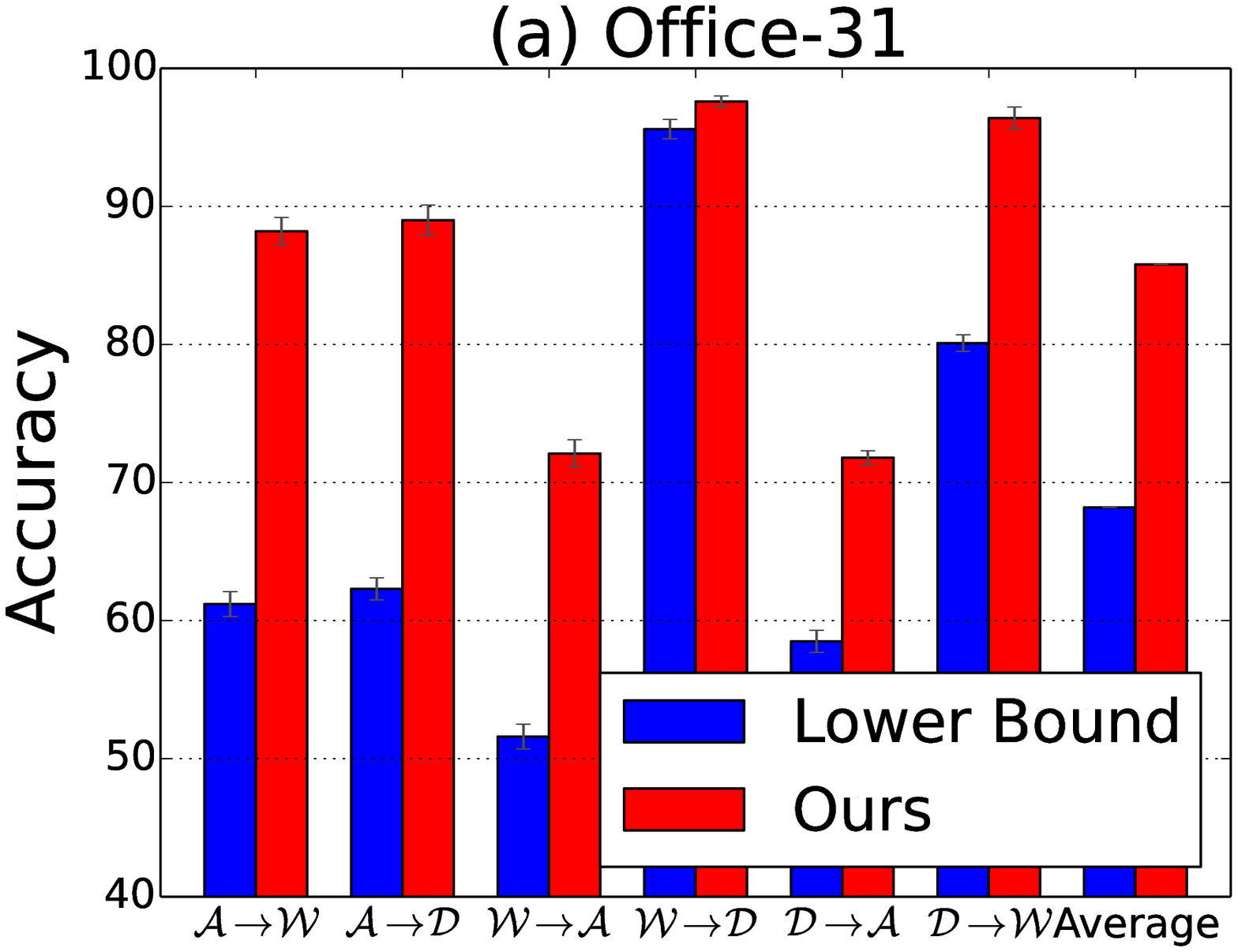}
\includegraphics[width=0.48\linewidth]{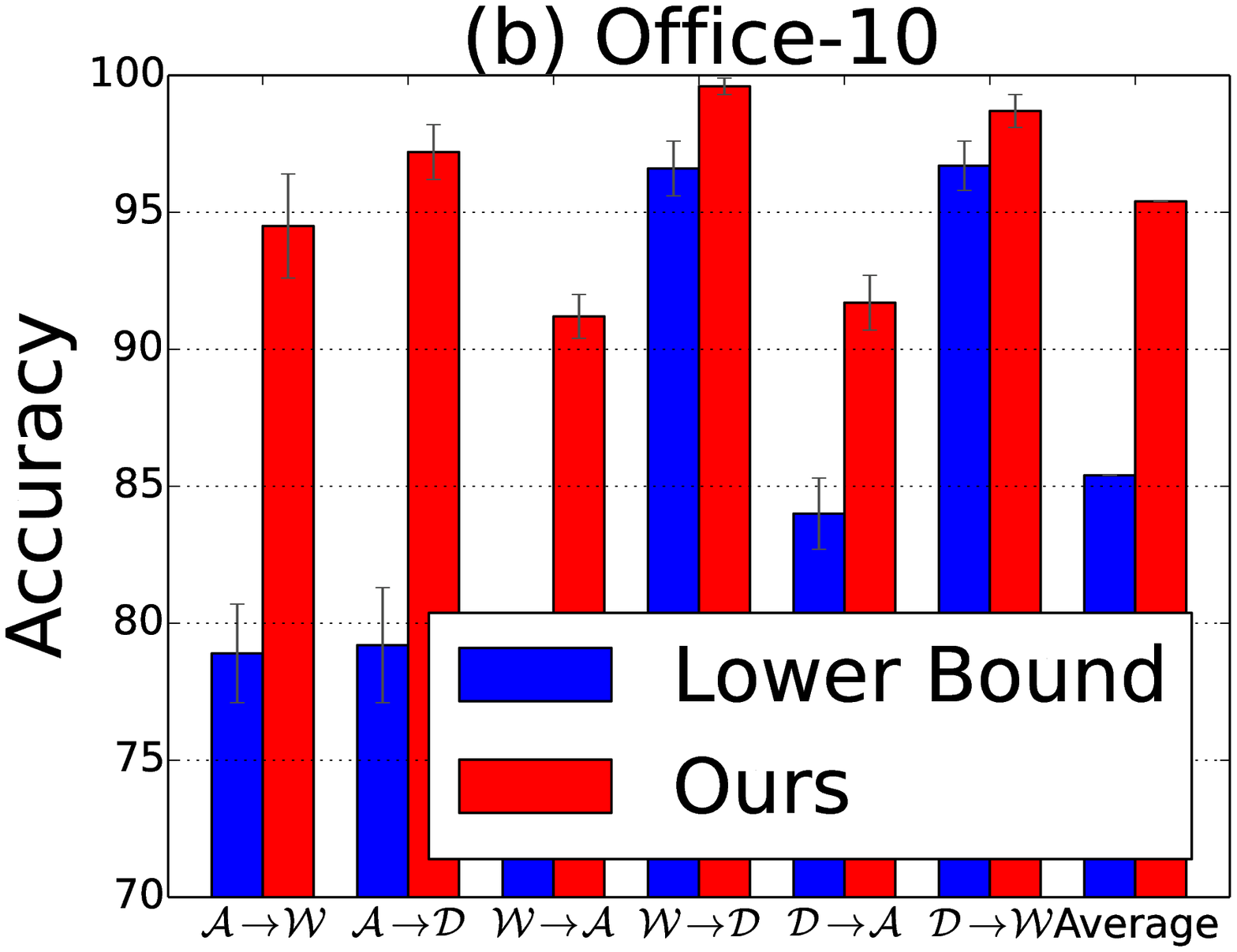} \\
\includegraphics[width=0.48\linewidth]{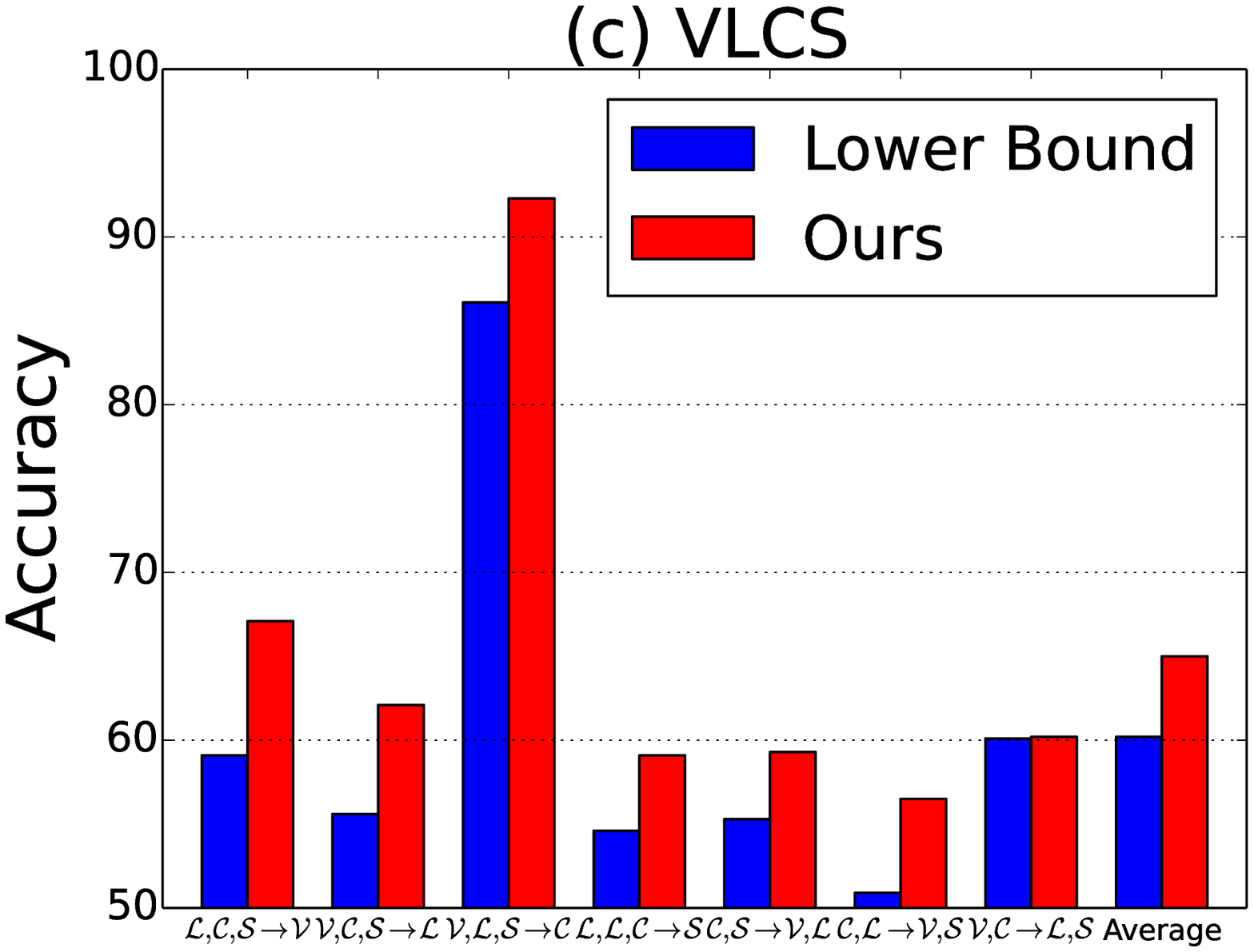}
\includegraphics[width=0.48\linewidth]{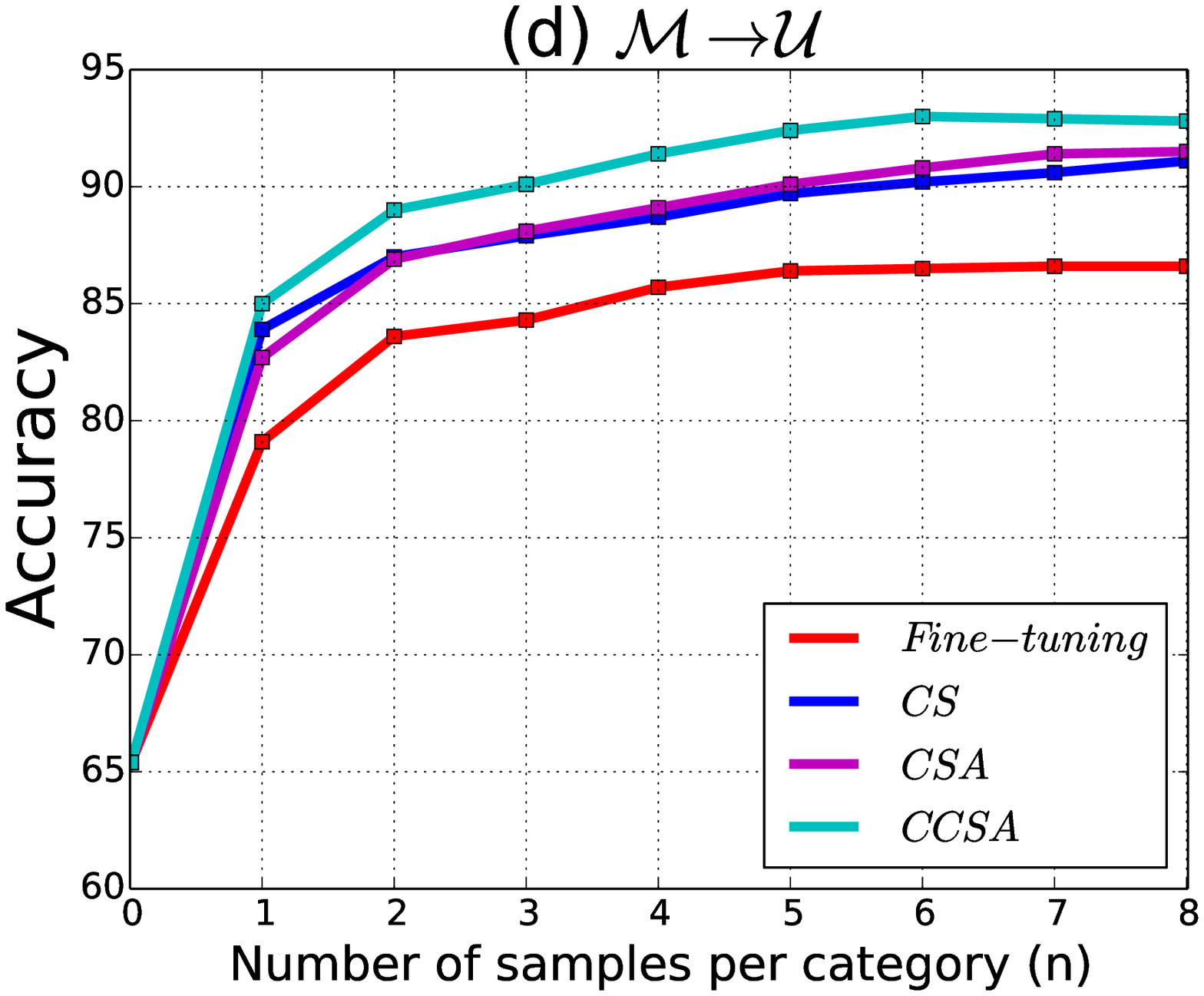}
\end{center}
\caption{\textbf{(a), (b), (c):} Improvement of \modelDA over the base model. \textbf{(d):} Average classification accuracy for $\mathcal{M} \rightarrow \mathcal{U}$ task for different number of labeled target samples per category ($n$). It shows that our model provides significant improvement over baselines.}
\label{fig-plots}
\end{figure}

\section{Extension to Domain Generalization} \label{sec-Deep-DG}

In visual domain generalization (DG), $D$ labeled datasets $\mathcal{D}_{s_1}$, $\cdots$, $\mathcal{D}_{s_D}$, representative of $D$ distinct source domains are given. The goal is to learn from them a visual classifier $f$ that during testing is going to perform well on data $\mathcal{D}_t$, not available during training, thus representative of an unknown target domain.

The SDA method in Section~\ref{sec-DA} treats source and target datasets $\mathcal{D}_s$  and $\mathcal{D}_t$ almost symmetrically. In particular, the embedding $g$ aims at achieving semantic alignment, while favoring class separation. The only asymmetry is in the prediction function $h$ that is trained only on the source, to be then fine-tuned on the target.

In domain generalization, we are not interested in adapting the classifier to the target domain, because it is unknown. Instead, we want to make sure that the embedding $g$ maps to a domain invariant space. To do so we consider every distinct unordered pair of source domains $(u,v)$, represented by $\mathcal{D}_{s_u}$ and $\mathcal{D}_{s_v}$, and, like in SDA, impose the semantic alignment loss~\eqref{eq-sda-semantic-confusion} as well as the separation loss~\eqref{eq-sda-separation}. Moreover, the losses are summed over every pair in order to make the map $g$ as domain invariant as possible. Similarly, the classifier $h$ should be as correct as possible for any of the mapped samples, to maximize performance on an unseen target. This calls for having a fully symmetric learning for $h$ by training it on all the source domains, meaning that the classification loss~\eqref{eq-basic} is summed over every domain $s_u$. See Figure~\ref{fig-domain_generalization}.

The network architecture is still the one in Figure~\ref{fig-model}, and we have implemented it with the same choices for distances and similarities as those made in Section~\ref{sec-Deep-DA}. However, since we are summing the losses~\eqref{eq-sda-semantic-confusion} and~\eqref{eq-sda-separation} over every unordered pair of source domains, there is a quadratic growth of paired training samples. So, if necessary, rather than processing every paired sample, we select them randomly.


\section{Experiments} \label{sec-Experiments}

\begin{table*}[t]
\caption{\textbf{Office dataset.} Classification accuracy for domain adaptation over the 31 categories of the Office dataset. $\mathcal{A}$, $\mathcal{W}$, and $\mathcal{D}$ stand for Amazon, Webcam, and DSLR domain. {\tt Lower Bound} is our base model without adaptation.}
\label{tab-Office-all-Classes}
\centering
\resizebox{1.5\columnwidth}{!}{%
  \begin{tabular}{ | l | c | c | c | c | c | c | c |}
\cline{3-8}
   \multicolumn{2}{c}{\tt{}} & \multicolumn{3}{|c|}{\tt{Unsupervised}} & \multicolumn{3}{|c|}{\tt{Supervised}}\\
\hline
            & {\tt Lower Bound} & {\tt ~\cite{tzeng2014deep}} & {\tt ~\cite{long2015learning}} & {\tt ~\cite{ghifary2016deep}} &  {\tt ~\cite{tzengHDS15iccv}} &{\tt ~\cite{koniusz2016domain}} & {\tt \modelDA}  \\
\hline
{\sl $\mathcal{A} \rightarrow \mathcal{W}$}  & 61.2 $\pm$ 0.9 & 61.8 $\pm$ 0.4 & 68.5 $\pm$ 0.4 & 68.7 $\pm$ 0.3   &  82.7 $\pm$ 0.8 & 84.5 $\pm$ 1.7  &  {\bf 88.2 $\pm$ 1.0}   \\
{\sl $\mathcal{A} \rightarrow \mathcal{D}$}  & 62.3 $\pm$ 0.8 & 64.4 $\pm$ 0.3 & 67.0 $\pm$ 0.4 & 67.1 $\pm$ 0.3   & 86.1 $\pm$ 1.2 & 86.3 $\pm$ 0.8  & {\bf 89.0 $\pm$ 1.2}    \\
{\sl $\mathcal{W} \rightarrow \mathcal{A}$}  & 51.6 $\pm$ 0.9 & 52.2 $\pm$ 0.4 & 53.1 $\pm$ 0.3 & 54.09 $\pm$ 0.5  & 65.0 $\pm$ 0.5 & 65.7 $\pm$ 1.7 &  {\bf 72.1 $\pm$ 1.0}   \\
{\sl $\mathcal{W} \rightarrow \mathcal{D}$}  & 95.6 $\pm$ 0.7  & 98.5 $\pm$ 0.4 & {\bf 99.0 $\pm$ 0.2} & {\bf 99.0 $\pm$ 0.2}   & 97.6 $\pm$ 0.2 & 97.5 $\pm$ 0.7  &  97.6 $\pm$ 0.4   \\
{\sl $\mathcal{D} \rightarrow \mathcal{A}$}  & 58.5 $\pm$ 0.8  & 52.1 $\pm$ 0.8 & 54.0 $\pm$ 0.4 & 56.0 $\pm$ 0.5 &  66.2 $\pm$ 0.3   & 66.5 $\pm$ 1.0 &  {\bf 71.8 $\pm$ 0.5}   \\
{\sl $\mathcal{D} \rightarrow \mathcal{W}$}  & 80.1 $\pm$ 0.6 & 95.0 $\pm$ 0.5 & 96.0 $\pm$ 0.3 & {\bf 96.4 $\pm$ 0.3}   & 95.7 $\pm$ 0.5 & 95.5 $\pm$ 0.6 & {\bf 96.4 $\pm$ 0.8}  \\
\hline
{\sl Average}  & 68.2 & 70.6 & 72.9 & 73.6   & 82.21 & 82.68 & {\bf 85.8}  \\
\hline
    \end{tabular}}
\end{table*}

\begin{table}[t]
\caption{\textbf{Office dataset.} Classification accuracy for domain adaptation over the Office dataset when only the labeled target samples of 15 classes are available during training. Testing is done on all 31 classes. $\mathcal{A}$, $\mathcal{W}$, and $\mathcal{D}$ stand for Amazon, Webcam, and DSLR domain. {\tt Lower Bound} is our base model without adaptation.}
\label{tab-Office-missing-Classes}
\centering
\resizebox{.7\columnwidth}{!}{%
  \begin{tabular}{ | l | c | c | c |}
\hline
            & {\tt Lower Bound} & {\tt ~\cite{tzengHDS15iccv}} & {\tt \modelDA}  \\
\hline
{\sl $\mathcal{A} \rightarrow \mathcal{W}$}  & 52.1 $\pm$ 0.6 & 59.3 $\pm$ 0.6 & {\bf 63.3 $\pm$ 0.9} \\
{\sl $\mathcal{A} \rightarrow \mathcal{D}$}  & 61.6 $\pm$ 0.8 & 68.0 $\pm$ 0.5 & {\bf 70.5 $\pm$ 0.6} \\
{\sl $\mathcal{W} \rightarrow \mathcal{A}$}  & 34.5 $\pm$ 0.9 & 40.5 $\pm$ 0.2 & {\bf 43.6 $\pm$ 1.0}  \\
{\sl $\mathcal{W} \rightarrow \mathcal{D}$}  & 95.1 $\pm$ 0.2 & {\bf 97.5 $\pm$ 0.1} & 96.2 $\pm$ 0.3   \\
{\sl $\mathcal{D} \rightarrow \mathcal{A}$}  & 40.1 $\pm$ 0.3 & {\bf 43.1 $\pm$ 0.2} & 42.6 $\pm$ 0.6  \\
{\sl $\mathcal{D} \rightarrow \mathcal{W}$}  & 89.7 $\pm$ 0.8 & {\bf 90.0 $\pm$ 0.2} & {\bf 90.0 $\pm$ 0.2}  \\
\hline
{\sl Average}  & 62.26 & 66.4 & 67.83 \\
\hline
    \end{tabular}}
\end{table}

\begin{table*}[t]
\caption{\textbf{Office dataset.} Classification accuracy for domain adaptation over the 10 categories of the Office dataset. $\mathcal{A}$, $\mathcal{W}$, and $\mathcal{D}$ stand for Amazon, Webcam, and DSLR domain. {\tt Lower Bound} is our base model with no adaptation.}
\label{tab-Office-10-Classes}
\centering
\resizebox{1.4\columnwidth}{!}{%
  \begin{tabular}{| l | c | c | c | c | c | c|}
\cline{2-7}
         \multicolumn{1}{c|}{\tt{}}   & {\tt Lower Bound} & {\tt GFK~\cite{gong2012geodesic}} & {\tt mSDA~\cite{chen2012marginalized}} & {\tt CDML~\cite{wang2014cross}} &  {\tt RTML~\cite{ding2017robust}} & {\tt \modelDA}  \\
\cline{2-7}
 \multicolumn{1}{c|}{\tt{}} & \multicolumn{6}{|c|}{\bf{SURF}}\\
\hline
{\sl $\mathcal{A} \rightarrow \mathcal{W}$} & 26.5 $\pm$ 3.1  & 39.9 $\pm$ 0.9 & 35.5 $\pm$ 0.5 & 37.3 $\pm$ 0.7 & 43.4 $\pm$ 0.9   &  {\bf 71.2 $\pm$ 1.3}    \\
{\sl $\mathcal{A} \rightarrow \mathcal{D}$} & 17.5 $\pm$ 1.2  & 36.2 $\pm$ 0.7 & 29.7 $\pm$ 0.7 & 35.3 $\pm$ 0.5 & 43.3 $\pm$ 0.6   &  {\bf 74.2 $\pm$ 1.3}    \\
{\sl $\mathcal{W} \rightarrow \mathcal{A}$} & 25.9 $\pm$ 1.0  & 29.8 $\pm$ 0.6 & 32.1 $\pm$ 0.8 & 32.4 $\pm$ 0.5 & 37.5 $\pm$ 0.7   &  {\bf 42.9 $\pm$ 0.9}    \\
{\sl $\mathcal{W} \rightarrow \mathcal{D}$} & 46.9 $\pm$ 1.1  & 80.9 $\pm$ 0.4 & 56.6 $\pm$ 0.4 & 77.9 $\pm$ 0.9 & {\bf 91.7 $\pm$ 1.1}   &  85.1 $\pm$ 1.0    \\
{\sl $\mathcal{D} \rightarrow \mathcal{A}$} & 19.3 $\pm$ 1.9  & 33.2 $\pm$ 0.6 & 33.6 $\pm$ 0.8 & 29.4 $\pm$ 0.8 & {\bf 36.3 $\pm$ 0.3}   &  28.9 $\pm$ 1.3    \\
{\sl $\mathcal{D} \rightarrow \mathcal{W}$} & 48.0 $\pm$ 2.1  & 79.4 $\pm$ 0.6 & 68.6 $\pm$ 0.7 & 79.4 $\pm$ 0.6 & {\bf 90.5 $\pm$ 0.7}   &  77.3 $\pm$ 1.6    \\
\hline
{\sl Average}  & 30.6 & 43.5 & 38.4 & 43.5 & 49.8   & {\bf 63.2}   \\
\hline
 \multicolumn{1}{c|}{\tt{}} & \multicolumn{6}{|c|}{\bf{DeCaF-fc6}}\\
\hline
{\sl $\mathcal{A} \rightarrow \mathcal{W}$} & 78.9 $\pm$ 1.8  & 73.1 $\pm$ 2.8 & 64.6 $\pm$ 4.2 & 75.9 $\pm$ 2.1 & 79.5 $\pm$ 2.6   &  {\bf 94.5 $\pm$ 1.9}    \\
{\sl $\mathcal{A} \rightarrow \mathcal{D}$} & 79.2 $\pm$ 2.1 & 82.6 $\pm$ 2.1 & 72.6 $\pm$ 3.5 & 81.4 $\pm$ 2.6 & 83.8 $\pm$ 1.7    & {\bf 97.2 $\pm$ 1.0}   \\
{\sl $\mathcal{W} \rightarrow \mathcal{A}$} & 77.3 $\pm$ 1.1 & 82.6 $\pm$ 1.3 & 71.4 $\pm$ 1.7 & 86.3 $\pm$ 1.6 & 90.8 $\pm$ 1.6   & {\bf 91.2 $\pm$ 0.8}   \\
{\sl $\mathcal{W} \rightarrow \mathcal{D}$} & 96.6 $\pm$ 1.0 & 98.8 $\pm$ 0.9  & 99.5 $\pm$ 0.6 & 99.4 $\pm$ 0.4 & {\bf 100 $\pm$ 0.0}   & 99.6 $\pm$ 0.5   \\
{\sl $\mathcal{D} \rightarrow \mathcal{A}$} & 84.0 $\pm$ 1.3 & 85.4 $\pm$ 0.7  & 78.8 $\pm$ 0.5 & 88.4 $\pm$ 0.5 & 90.6 $\pm$ 0.5  & {\bf 91.7 $\pm$ 1.0}    \\
{\sl $\mathcal{D} \rightarrow \mathcal{W}$} & 96.7 $\pm$ 0.9  & 91.3 $\pm$ 0.4 & 97.5 $\pm$ 0.4 & 95.1 $\pm$ 0.5 & 98.6 $\pm$ 0.3  & {\bf 98.7 $\pm$ 0.6}   \\
\hline
{\sl Average}  & 85.4 & 85.63 & 80.73 & 87.75 & 90.55   & {\bf 95.4}   \\
\hline
    \end{tabular}}
\end{table*}

\begin{table}[t]
\caption{\textbf{MNIST-USPS datasets.} Classification accuracy for domain adaptation over the MNIST and USPS datasets. $\mathcal{M}$ and $\mathcal{U}$ stand for MNIST and USPS domain. {\tt Lower Bound} is our base model without adaptation. {\tt \modelDA- $n$} stands for our method when we use $n$ labeled target samples per category in training.}
\label{tab-MNIST-USPS}
\centering
\resizebox{.65\columnwidth}{!}{%
  \begin{tabular}{ | l | c | c | c |}
\hline
    Method        & $\mathcal{M} \rightarrow \mathcal{U}$ & $\mathcal{U} \rightarrow \mathcal{M}$ & {\sl Average} \\
\hline
{\tt ADDA~\cite{Tzeng_2017_CVPR}}   & 89.4  & 90.1 & 89.7 \\
{\tt CoGAN~\cite{liu2016coupled}}   & 91.2  & 89.1 & 90.1 \\
\hline
{\tt Lower Bound}                & 65.4  & 58.6 & 62.0 \\
{\tt CCSA-1}                     & 85.0  & 78.4 & 81.7 \\
{\tt CCSA-2}                     & 89.0  & 82.0 & 85.5 \\
{\tt CCSA-3}                     & 90.1  & 85.8 & 87.9 \\
{\tt CCSA-4}                     & 91.4  & 86.1 & 88.7 \\
{\tt CCSA-5}                     & 92.4  & 88.8 & 90.1 \\
{\tt CCSA-6}                     & 93.0  & 89.6 & 91.3 \\
{\tt CCSA-7}                     & 92.9  & 89.4 & 91.1 \\
{\tt CCSA-8}                     & 92.8  & 90.0 & 91.4 \\
\hline
    \end{tabular}}
\end{table}

We divide the experiments into two parts, domain adaptation and domain generalization. In both sections, we use
benchmark datasets and compare our domain adaptation
model and our domain generalization model, both indicated
as CCSA, with the state-of-the-art.

\subsection{Domain Adaptation}

\begin{table}[t]
\caption{\textbf{VLCS dataset.} Classification accuracy for domain generalization over the 5 categories of the VLCS dataset. {\tt LB} (Lower Bound) is our base model trained without the contrastive semantic alignment loss. {\tt 1NN} stands for first nearest neighbor. }
\label{tab-VLCS}
\centering
\resizebox{0.92\columnwidth}{!}{%
  \begin{tabular}{ | l | c | c | c | c | c | c | c |}
\cline{2-8}
   \multicolumn{1}{c}{\tt{}} & \multicolumn{3}{|c|}{\tt{Lower Bound}} & \multicolumn{4}{|c|}{\tt{Domain Generalization}}\\
\hline
            & {\tt 1NN} & {\tt SVM} & {\tt LB} & {\tt UML~\cite{FXRQ_iccv13}} &  {\tt {\small LRE-SVM~\cite{xuLNX14eccv}}} &{\tt SCA~\cite{ghifary2016scatter}} & {\tt \modelGA}  \\
\hline
{\sl $\mathcal{L},\mathcal{C},\mathcal{S} \rightarrow \mathcal{V}$}  
& 57.2 & 58.4 & 59.1 & 56.2 &  60.5  & 64.3  & {\bf 67.1}   \\
{\sl $\mathcal{V},\mathcal{C},\mathcal{S} \rightarrow \mathcal{L}$}
& 52.4 & 55.2 & 55.6 & 58.5 &  59.7  & 59.6  & {\bf 62.1}   \\
{\sl $\mathcal{V},\mathcal{L},\mathcal{S} \rightarrow \mathcal{C}$}
& 90.5 & 85.1 & 86.1 & 91.1 &  88.1  & 88.9  & {\bf 92.3}   \\
{\sl $\mathcal{V},\mathcal{L},\mathcal{C} \rightarrow \mathcal{S}$}
& 56.9 & 55.2 & 54.6 & 58.4 &  54.8  & {\bf 59.2}  & 59.1   \\
{\sl $\mathcal{C},\mathcal{S} \rightarrow \mathcal{V},\mathcal{L}$}
& 55.0 & 55.5 & 55.3 & 56.4 &  55.0  & {\bf 59.5}  & 59.3   \\
{\sl $\mathcal{C},\mathcal{L} \rightarrow \mathcal{V},\mathcal{S}$}
& 52.6 & 51.8 & 50.9 & {\bf 57.4} &  52.8  & 55.9  & 56.5   \\
{\sl $\mathcal{V},\mathcal{C} \rightarrow \mathcal{L}, \mathcal{S}$}
& 56.6 & 59.9 & 60.1 & 55.4 &  58.8  & {\bf 60.7}  & 60.2   \\
\hline
{\sl Average}  
& 60.1 & 60.1 & 60.2 & 61.5 &  61.4  & 64.0 & {\bf 65.0}  \\
\hline
    \end{tabular}}
\end{table}

We present results using the Office dataset~\cite{saenkoKFD2010eccv}, the MNIST dataset~\cite{lecun1998gradient}, and the USPS dataset~\cite{hull1994database}.

\subsubsection{Office Dataset} 

The office dataset is a standard benchmark dataset for visual domain adaptation. It contains 31 object classes for three domains: Amazon, Webcam, and DSLR, indicated as $\mathcal{A}$, $\mathcal{W}$, and $\mathcal{D}$, for a total of 4,652 images. We consider six domain shifts using the three domains ($\mathcal{A} \rightarrow \mathcal{W}$, $\mathcal{A} \rightarrow \mathcal{D}$, $\mathcal{W} \rightarrow \mathcal{A}$, $\mathcal{W} \rightarrow \mathcal{D}$, $\mathcal{D} \rightarrow \mathcal{A}$, and $\mathcal{D} \rightarrow \mathcal{W}$). We performed different experiments using this dataset.

\noindent \textbf{First experiment.} We followed the setting described in~\cite{tzengHDS15iccv}. All classes of the office dataset and 5 train-test splits are considered. For the source domain, 20 examples per category for the Amazon domain, and 8 examples per category for the DSLR and Webcam domains are randomly selected for training for each split. Also, 3 labeled examples are randomly selected for each category in the
target domain for training for each split. The rest of the target samples are used for testing. Note that we used the same splits generated by~\cite{tzengHDS15iccv}. We also report the classification results of the SDA algorithm presented in~\cite{long2015learning} and~\cite{koniusz2016domain}. In addition to the SDA algorithms, we report the results of some recent UDA algorithms. They follow a different experimental protocol compared to the SDA algorithms, and use all samples of the target domain in training as unlabeled data together with all samples of the source domain. 

For the embedding function $g$, we used the convolutional layers of the VGG-16 architecture~\cite{Simonyan14c} followed by 2 fully connected layers with output size of 1024 and 128, respectively. For the prediction function $h$, we used a fully connected layer with softmax activation. Similar to~\cite{tzengHDS15iccv}, we used the weights pre-trained on the ImageNet dataset~\cite{imagenet2015} for the convolutional layers, and initialized the fully connected layers using all the source domain data. We then fine-tuned all the weights using the train-test splits.

Table~\ref{tab-Office-all-Classes} reports the classification accuracy over $31$ classes for the Office dataset and shows that \modelDA has better performance compared to~\cite{tzengHDS15iccv}. Since the difference between $\mathcal{W}$ domain and $\mathcal{D}$ domain is not considerable, unsupervised algorithms work well on $\mathcal{D} \rightarrow \mathcal{W}$ and $\mathcal{W} \rightarrow \mathcal{D}$.   However, in the cases when target and source domains are very different ($\mathcal{A} \rightarrow \mathcal{W}$, $\mathcal{W} \rightarrow \mathcal{A}$, $\mathcal{A} \rightarrow \mathcal{D}$, and $\mathcal{D} \rightarrow \mathcal{A}$), \modelDA shows larger margins compared to the second best. This suggests that \modelDA will provide greater alignment gains when there are bigger domain shifts. Figure~\ref{fig-plots}(a) instead, shows how much improvement can be obtained with respect to the base model. This is simply obtained by training $g$ and $h$ with only the classification loss and source training data, so no adaptation is performed.

\noindent \textbf{Second experiment.}  We followed the setting described in~\cite{tzengHDS15iccv} when only 10 target labeled samples of 15 classes of the Office dataset are available during training. Similar to~\cite{tzengHDS15iccv}, we compute the accuracy on the remaining 16 categories for which no target data was available during training. We used the same network structure as in the first experiment and the same splits generated by~\cite{tzengHDS15iccv}.

Table~\ref{tab-Office-missing-Classes} shows that \modelDA is effective at transferring information from the labeled classes to the unlabeled target classes. Similar to the first experiment, \modelDA works well when shifts between domains are larger.

\noindent \textbf{Third experiment.} We used the original train-test splits of the Office dataset~\cite{saenkoKFD2010eccv}. The splits are generated in a similar manner to the first experiment but here instead, only 10 classes are considered (backpack, bike, calculator, headphones, keyboard, laptop-computer, monitor, mouse, mug, and projector). In order to compare our results with the state-of-the-art, we used DeCaF-fc6 features~\cite{Donahue13decaf} and 800-dimension SURF features as input.  For DeCaF-fc6 features (SURF features) we used 2 fully connected layers with output size of 1024 (512) and 128 (32) with ReLU activation as the embedding function, and one fully connected layer with softmax activation as the prediction function.
The features and splits are available on the Office dataset webpage~\footnote{\url{https://cs.stanford.edu/~jhoffman/domainadapt/}}.

We compared our results with three UDA (GFK~\cite{gong2012geodesic}, mSDA~\cite{chen2012marginalized}, and RTML~\cite{ding2017robust}) and one SDA (CDML~\cite{wang2014cross}) algorithms under the same settings. Table~\ref{tab-Office-10-Classes} shows that \modelDA provides an improved accuracy with respect to the others. Again, greater domain shifts are better compensated by \modelDA. Figure~\ref{fig-plots}(b) shows the improvement of \modelDA over the base model using DeCaF-fc6 features.

\subsubsection{MNIST-USPS Datasets} \label{sec-MNIST-USPS}

The MNIST ($\mathcal{M}$) and USPS ($\mathcal{U}$) datasets have recently been used for domain adaptation~\cite{fernandoTT15prl,rozantsev2016beyond}. They contain images of digits from 0 to 9. We considered two cross-domain tasks, $\mathcal{M} \rightarrow \mathcal{U}$ and $\mathcal{U} \rightarrow \mathcal{M}$, and followed the experimental setting in~\cite{fernandoTT15prl,rozantsev2016beyond}, which involves randomly selecting 2000 images from MNIST and 1800 images from USPS. Here, we randomly selected $n$ labeled samples per class from target domain data and used them in training. We evaluated our approach for $n$ ranging from 1 to 8 and repeated each experiment $10$ times (we only show the mean of the accuracies because the standard deviation is very small).

Similar to~\cite{lecun1998gradient}, we used $2$ convolutional layers with 6 and 16 filters of $5\times5$ kernels followed by max-pooling layers and $2$ fully connected layers with size $120$ and $84$ as the embedding function $g$, and one fully connected layer with softmax activation as the prediction function $h$. We compare our method with $2$ recent UDA methods. Those methods use all target samples in their training stage, while we only use very few labeled target samples per category in training.

Table~\ref{tab-MNIST-USPS} shows the average classification accuracy of the MNIST-USPS datasets. \modelDA works well even when only one target sample per category ($n=1$) is available in training. Also, we can see that by increasing $n$, the accuracy quickly converges to the top.

\noindent \textbf{Ablation study.} We consider three baselines to compare with \modelDA  for $\mathcal{M} \rightarrow \mathcal{U}$ task. First, we train the network with source data and then fine-tune it with available target data. Second, we train the network using the classification and semantic
alignment losses ($\mathcal{L}_{CSA} (f) = \mathcal{L}_{C} (h\circ g) + \mathcal{L}_{SA} (g)$). Third, we train the network using the classification and separation losses ($\mathcal{L}_{CS} (f) = \mathcal{L}_{C} (h\circ g) + \mathcal{L}_{S} (g)$). Figure~\ref{fig-plots}(d) shows the average accuracies over $10$ repetitions. It shows that CSA and CS improve the accuracy over fine-tuning. Using the semantic alignment loss together with separation loss (CCSA) shows the best performance.

\noindent \textbf{Visualization.} We show how samples lie on the embedding space using \modelDA. First, we considered the row images of the MNIST and USPS datasets and plotted 2D visualization of them using t-SNE algorithm~\cite{maaten2008visualizing}. As Figure~\ref{fig-domain_embeded}(Left) shows the row images of the same class and different domains lie far away from each other in the 2D subspace. For example, the samples of the class zero of the USPS dataset ($0\; U$) are far from the class zero of the MNIST dataset ($0\; M$). Second, we trained our base model with no adaptation on the MNIST dataset. We then plotted the 2D visualization of the MNIST and USPS samples in the embedding space (output of $g$, the last fully connected layer). As Figure~\ref{fig-domain_embeded}(Middle) shows, the samples from the same class and different domains still lie far away from each other in the 2D subspace. Finally, we trained our SDA model on the MNIST dataset and $3$ labeled samples per class of the USPS dataset. We then plotted the 2D visualization of the MNIST and USPS samples in the embedding space (output of $g$). As Figure~\ref{fig-domain_embeded}(Right) shows, the samples from the same class and different domains now lie very close to each other in the 2D subspace. Note however, that this is only a 2D visualization of high-dimensional data, and Figure~\ref{fig-domain_embeded}(Right) may not perfectly reflect how close is the data from the same class, and how classes are separated.

\noindent \textbf{Weight sharing}: There is no restriction whether $g_t$ and $g_s$ should share the weights or not. Not sharing weights will likely lead to overfitting, given the reduced amount of target training data, and weight-sharing can be seen as a regularizer. We repeated the experiment for $\mathcal{M} \rightarrow \mathcal{U}$ task when $n=4$. Not sharing weights provides the average accuracy $88.6$ over $10$ repetitions  which is less than the average accuracy with weight-sharing (see Table~\ref{tab-MNIST-USPS}). A similar behavior can be seen for other experiments.

\subsection{Domain Generalization}

We evaluate \modelGA on different datasets. The goal is to show that \modelGA is able to learn a domain invariant embedding subspace for visual recognition tasks.

\subsection{VLCS Dataset}
In this section, we use images of $5$ shared object categories (bird, car, chair, dog, and person), of the PASCAL VOC2007 ($\mathcal{V}$)~\cite{everingham2010pascal}, LabelMe ($\mathcal{L}$)~\cite{russell2008labelme},
Caltech-101 ($\mathcal{C}$)~\cite{fei2007learning}, and SUN09 ($\mathcal{S}$)~\cite{choi2010exploiting} datasets, which is known as VLCS dataset~\cite{FXRQ_iccv13}.

\cite{ghifary2015domain,ghifary2016scatter} have shown that there are covariate shifts between the above $4$ domains and have developed a DG method to minimize them. We followed their experimental setting, and randomly divided each domain into a training set ($70\%$) and a test set ($30\%$) and conducted a {\tt leave-one-domain-out} evaluation ($4$ cross-domain cases) and a {\tt leave-two-domain-out} evaluation ($3$ cross-domain cases). In order to compare our results with the state-of-the-art, we used DeCaF-fc6 features which are publicly available~\footnote{\url{http://www.cs.dartmouth.edu/~chenfang/proj_page/FXR_iccv13/index.php}}, and repeated each cross-domain case $20$ times and reported the average classification accuracy.

We used $2$ fully connected layers with output size of $1024$ and $128$ with ReLU activation as the embedding function $g$, and one fully connected layer with softmax activation as the prediction function $h$.
To create positive and negative pairs for training our network, for each sample of a source domain we randomly selected 5 samples from each remaining source domain, and help in this way to avoid overfitting. However, to train a deeper network together with convolutional layers, it is enough to create a large amount of positive and negative pairs.

We report comparative results in Table~\ref{tab-VLCS}, where all DG methods work better than the base model, emphasizing the need for domain generalization. Our DG method has higher average performance. Also, note that in order to compare with the state-of-the-art DG methods, we only used 2 fully connected layers for our network and precomputed features as input. However, when using convolutional layers on row images, we expect our DG model to provide better performance. Figure~\ref{fig-plots}(c) shows the improvement of our DG model over the base model using DeCaF-fc6 features.

\subsection{MNIST Dataset}
We followed the setting in~\cite{ghifary2015domain}, and randomly selected a set $M$ of $100$ images per category from the MNIST dataset ($1000$ in total). We then rotated each image in $M$ five times with $15$ degrees intervals, creating five new domains $M_{15^{\circ}}$, $M_{30^{\circ}}$, $M_{45^{\circ}}$, $M_{60^{\circ}}$, and $M_{75^{\circ}}$. We conducted a {\tt leave-one-domain-out} evaluation ($6$ cross-domain cases in total). We used the same network of Section~\ref{sec-MNIST-USPS}, and we repeated the experiments $10$ times. To create positive and
negative pairs for training our network, for each sample of a
source domain we randomly selected 2 samples from each
remaining source domain. We report comparative average accuracies for \modelGA and others in Table~\ref{tab-MNIST}, showing again a performance improvement.

\begin{table}[t]
\caption{\textbf{MNIST dataset.} Classification accuracy for domain generalization over the MNIST dataset and its rotated domains. }
\label{tab-MNIST}
\centering
\resizebox{.9\columnwidth}{!}{%
  \begin{tabular}{ | c | c | c | c |}
\hline
         & {\sl CAE~\cite{rifai2011contractive}} & {\sl MTAE~\cite{ghifary2015domain}} & {\sl \modelGA} \\
\hline
$M_{15^{\circ}},M_{30^{\circ}},M_{45^{\circ}},M_{60^{\circ}},M_{75^{\circ}}  \rightarrow M $   
& 72.1  & 82.5 & {\bf 84.6} \\
$M,M_{30^{\circ}},M_{45^{\circ}},M_{60^{\circ}},M_{75^{\circ}}  \rightarrow M_{15^{\circ}} $           
& 95.3  & {\bf 96.3} & 95.6 \\
$M,M_{15^{\circ}},M_{45^{\circ}},M_{60^{\circ}},M_{75^{\circ}}  \rightarrow M_{30^{\circ}} $                 
& 92.6  & 93.4 & {\bf 94.6} \\
$M,M_{15^{\circ}},M_{30^{\circ}},M_{60^{\circ}},M_{75^{\circ}}  \rightarrow M_{45^{\circ}} $                      
& 81.5  & 78.6 & {\bf 82.9} \\
$M,M_{15^{\circ}},M_{30^{\circ}},M_{45^{\circ}},M_{75^{\circ}}  \rightarrow M_{60^{\circ}} $                      
& 92.7  & 94.2 & {\bf 94.8} \\
$M,M_{15^{\circ}},M_{30^{\circ}},M_{45^{\circ}},M_{60^{\circ}}  \rightarrow M_{75^{\circ}}  $                      
& 79.3  & 80.5 & {\bf 82.1} \\
\hline
Average 
& 85.5  & 87.5 & {\bf 89.1} \\
\hline
    \end{tabular}}
    
\end{table}


\vspace{-2mm}
\section{Conclusions}
\vspace{-2mm}
We have introduced a deep model in combination with the classification and contrastive semantic alignment (CCSA) loss to address SDA. We have shown that the CCSA loss can be augmented to address the DG problem without the need to change the basic model architecture. However, the approach is general in the sense that the architecture sub-components can be changed. We found that addressing the semantic distribution alignments with point-wise surrogates of distribution distances and similarities for SDA and DG works very effectively, even when labeled target samples are very few. In addition, we found the SDA accuracy to converge very quickly as more labeled target samples per category are available.
\vspace{-2mm}
\section*{Acknowledgments}
\vspace{-2mm}
{\small
This material is based upon work supported by the Center for Identification Technology Research and the National Science Foundation under Grant No. 1066197.}

{\small
\bibliographystyle{ieee}
\bibliography{iccv_DA}
}

\end{document}


\title{Supplementary Material: Unified Deep Supervised Domain Adaptation and Generalization}

\author{First Author\\
Institution1\\
Institution1 address\\
{\tt\small firstauthor@i1.org}
\and
Second Author\\
Institution2\\
First line of institution2 address\\
{\tt\small secondauthor@i2.org}
}

\maketitle

\section{Additional Experiments}

We add more results showing how the number of labeled target samples per class affects the performance of CCSAA, the proposed SDA approach. We use the same setting explained in Section 5.1.1 (first experiment) of the submitted paper, where we use 1, 2, and 3 labeled target samples per class during training. Moreover, $g$ is modeled by another network architecture, whereby the VGG-16~\cite{Simonyan14c} here is substituted by the convolutional layers of InceptionV3~\cite{szegedy2016rethinking}, followed by 2 fully connected layers with output size of 1024 and 128, respectively. For the prediction function $h$, we used a fully connected layer with softmax activation. We used the weights pre-trained on the ImageNet dataset for the convolutional layers, and we initialized the fully connected layers using all the source domain data. 

As explained in Section 3.2, after initialization, the learning is done end-to-end, where $f=h \circ g$ is trained by minimizing the loss function
\begin{equation}
\mathcal{L}_{CCSA} (f) = (1-\gamma)\mathcal{L}_{C} (h\circ g) + \gamma \left(\mathcal{L}_{SA} (g) + \mathcal{L}_{S} (g) \right) \; .
\label{eq-sda-contrastive-semantic-alignment}
\end{equation}
In this experiment, as well as in all those included in the submitted paper we set $\gamma = 0.1$. Also, in Eq. (10) of the paper, we set $m=1$. After the end-to-end learning, we use the training target labeled data to fine-tune the classification function $h$. Note that Eq. (7) and Eq. (8), are normalized averages and are implemented as such, while we have left out the normalization factors to keep the notation simple in the paper.

\begin{figure}[th!]
\begin{center}
\includegraphics[width=0.98\linewidth]{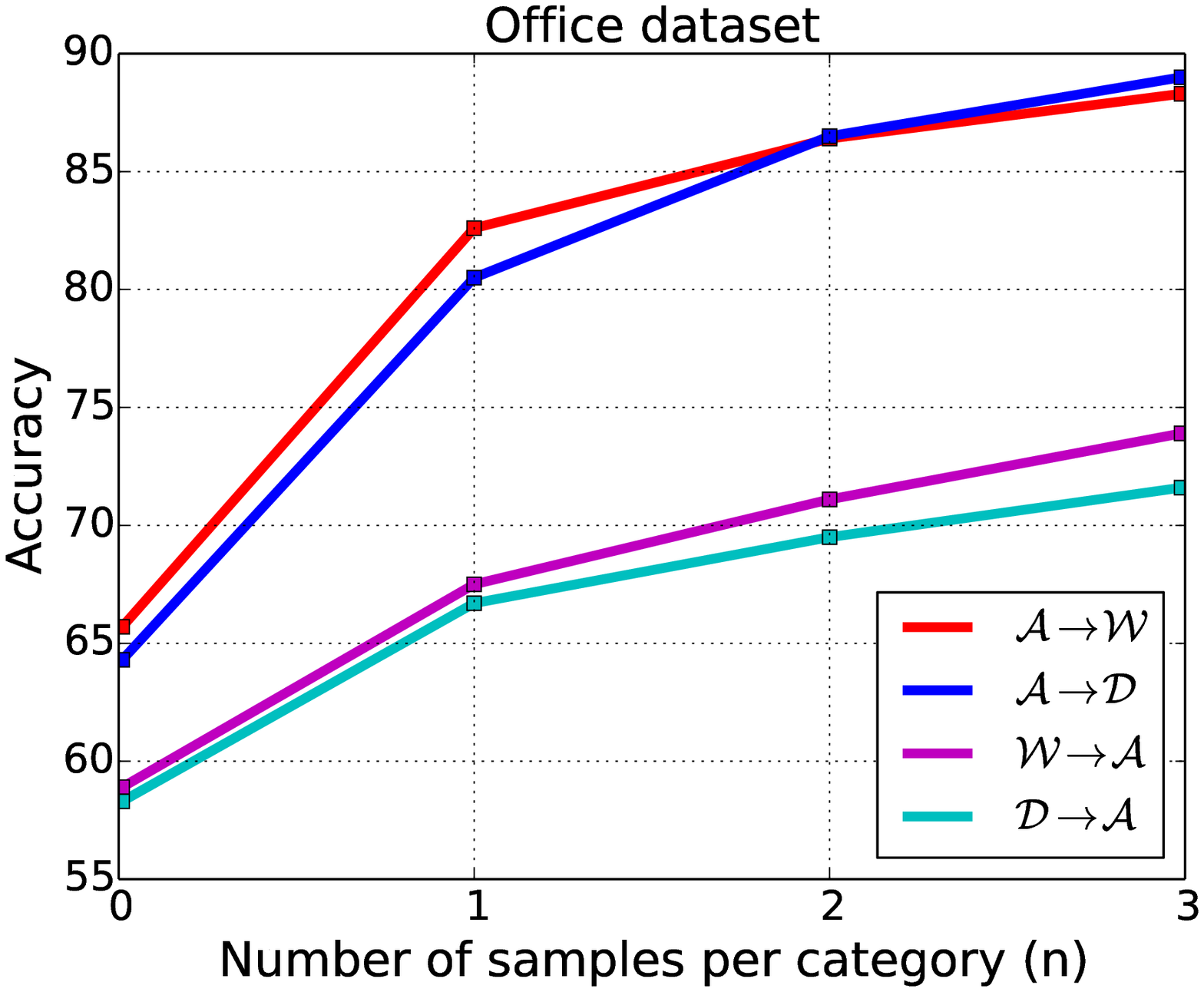}
\end{center}
\caption{\textbf{Office dataset.} Classification accuracy for domain adaptation. One labeled sample per category ($n=1$) increases the accuracy around 13$\%$ with respect to the {\tt Lower Bound} ($n=0$). This shows how quickly \modelDA semantically aligns domains as the number of labeled target training samples increase.}
\label{fig-office}
\end{figure}

Table~\ref{tab-Office-InceptionV3} shows the classification accuracy on the Office dataset, where {\tt \modelDA- $n$} indicates our method when we use $n$ labeled target samples per category in training.  CCSAA significantly improves the accuracy even when only one target sample per category ($n=1$) is available. Also, we see that increasing $n$ further increases the accuracy, but the margin gains decrease rather quickly, denoting a high speed of convergence towards the highest accuracy. See Figure~\ref{fig-office}.
\begin{table}[th!]
\caption{\textbf{Office dataset.} Classification accuracy. $\mathcal{A}$, $\mathcal{W}$, and $\mathcal{D}$ stand for Amazon, Webcam, and DSLR domain. {\tt Lower Bound} is our base model without adaptation.}
\label{tab-Office-InceptionV3}
\centering
\resizebox{.9\columnwidth}{!}{%
  \begin{tabular}{ | l | c | c | c | c |}
\hline
            & {\tt Lower Bound} & {\tt CCSAA - 1} & {\tt CCSAA - 2} & {\tt CCSAA - 3}  \\
\hline
{\sl $\mathcal{A} \rightarrow \mathcal{W}$}  & 65.7 $\pm$ 1.4 & 82.6 $\pm$ 0.3 & 86.4 $\pm$ 1.1 &  88.3 $\pm$ 0.8 \\
{\sl $\mathcal{A} \rightarrow \mathcal{D}$}  & 64.3 $\pm$ 0.9 & 80.5 $\pm$ 0.8 & 86.5 $\pm$ 0.5 &  89.0 $\pm$ 0.9 \\
{\sl $\mathcal{W} \rightarrow \mathcal{A}$}  & 58.9 $\pm$ 0.8 & 67.5 $\pm$ 0.3 & 71.1 $\pm$ 0.4 &  73.9 $\pm$ 1.0  \\
{\sl $\mathcal{W} \rightarrow \mathcal{D}$}  & 94.7 $\pm$ 0.6 & 97.6 $\pm$ 0.6 & 97.7 $\pm$ 0.7 &  98.0 $\pm$ 0.5  \\
{\sl $\mathcal{D} \rightarrow \mathcal{A}$}  & 58.3 $\pm$ 0.5 & 66.7 $\pm$ 0.9 & 69.5 $\pm$ 0.9 &  71.6 $\pm$ 1.0  \\
{\sl $\mathcal{D} \rightarrow \mathcal{W}$}  & 92.7 $\pm$ 0.3 & 94.8 $\pm$ 0.6 & 95.7 $\pm$ 0.6 &  96.2 $\pm$ 1.0  \\
\hline
{\sl Average}  & 72.4 & 81.6 & 84.4  & 86.1 \\
\hline
    \end{tabular}}
\end{table}

\subsection{Domain Generalization Implementation}

We add a few details on the implementation of domain generalization, originally left out because of space limitations. As explained in Section 4 of the paper, given $D$ labeled datasets $\mathcal{D}_{s_1}$, $\cdots$, $\mathcal{D}_{s_D}$, representative of $D$ distinct source domains, the end-to-end learning of $f = h \circ g$ is done by minimizing the loss function  \begin{equation}
\begin{split} 
\mathcal{L}_{CCSA} (f) & = 
\frac{1-\gamma}{D} \sum_{u} \mathcal{L}_{C} (h\circ g |  \mathcal{D}_{s_u}) \\ 
 +  \frac{2\gamma}{D^2 -D} \left( \sum_{(u,v)} \right. & \left.\mathcal{L}_{SA} (g | \mathcal{D}_{s_u}, \mathcal{D}_{s_v}) + \sum_{(u,v)} \mathcal{L}_{S}  (g | \mathcal{D}_{s_u},\mathcal{D}_{s_v}) \right)
\end{split}
\label{eq-DG}
\end{equation}
During training of the CCSAG model, because every unordered pair of domains $(u,v)$ is considered, we control the number of positive and negative training pairs by randomly picking a fraction of all the possible training pairs. For the VLCS dataset, for each sample of a source domain we randomly selected 5 samples from each remaining source domain, and also help in this way to avoid overfitting. For the MNIST dataset, for each sample of a source domain we randomly selected 2 samples from each remaining source domain. Similar to CCSAA, for all the experiments also for CCSAG we set $\gamma = 0.1$, and in Eq. (10) of the paper, we set $m=1$.

{\small
\bibliographystyle{ieee}
\bibliography{iccv_DA}
}